\documentclass[lettersize,journal]{IEEEtran}
\usepackage{cite}
\usepackage{amsmath, amsfonts, mathrsfs, amssymb}

\usepackage{amsthm}
\usepackage{algorithmic}
\usepackage{array}
\usepackage{textcomp}
\usepackage{stfloats}
\usepackage{multirow}
\usepackage{booktabs}       
\usepackage{nicefrac}       
\usepackage{microtype}      
\usepackage[caption=false,font=normalsize,labelfont=sf,textfont=sf]{subfig}
\usepackage{url}
\usepackage{verbatim}
\usepackage{cite}
\usepackage{graphicx}
\usepackage[switch]{lineno}
\usepackage[dvipsnames]{xcolor}
\usepackage[colorlinks=true, citecolor=Gray, linkcolor=OrangeRed, urlcolor=OrangeRed]{hyperref}
\usepackage{textcomp}
\def\BibTeX{{\rm B\kern-.05em{\sc i\kern-.025em b}\kern-.08em
    T\kern-.1667em\lower.7ex\hbox{E}\kern-.125emX}}

\usepackage{balance}
\newcommand{\etal}{\textit{et al}. }
\newcommand{\ie}{\textit{i}.\textit{e}., }
\newcommand{\eg}{\textit{e}.\textit{g}., }
\newtheorem{assumption}{Assumption}
\newtheorem{remark}{Remark}
\begin{document}
\title{Solving Energy-Independent Density for CT Metal Artifact Reduction via Neural Representation}
\author{Qing~Wu, Xu~Guo, Lixuan~Chen, Yanyan~Liu, Dongming~He, Xudong~Wang, Wuwei~Ren,\\Xueli~Chen,~\IEEEmembership{Senior Member, IEEE}, Yifeng~Zhang, S.~Kevin~Zhou,~\IEEEmembership{Fellow, IEEE},\\Jingyi~Yu,~\IEEEmembership{Fellow, IEEE}, and Yuyao~Zhang,~\IEEEmembership{Member, IEEE}

\thanks{This work was supported by the National Natural Science Foundation of China under Grants No. 62071299.}
\thanks{Qing Wu is with School of Information Science and Technology, ShanghaiTech University, Shanghai 201210, China, and with Shanghai Advanced Research Institute, Chinese Academy of Sciences, Shanghai 201210, China, and also with University of Chinese Academy of Sciences, Beijing 101408, China (e-mail: wuqing@shanghaitech.edu.cn).}
\thanks{Xu Guo and Yifeng Zhang are with School of Life Science and Technology, ShanghaiTech University, Shanghai 201210, China (e-mail: \{guoxu, zhangyf3\}@shanghaitech.edu.cn).}
\thanks{Lixuan Chen is with Electrical and Computer Engineering, University of Michigan, MI, 48105, United States (e-mail: chenlx@umich.edu).}
\thanks{Yanyan Liu is with Shanghai United Imaging Intelligence Co., Ltd., Shanghai, China (e-mail: yanyan.liu@united-imaging.com).}
\thanks{Dongming He and Xudong Wang are with Shanghai Ninth People's Hospital, Shanghai Jiao Tong University School of Medicine, Shanghai, China (e-mail: 1295227946@qq.com; xudongwang70@hotmail.com)}
\thanks{Xueli Chen is with Center for Biomedical-photonics and Molecular Imaging, Advanced Diagnostic-Therapy Technology and Equipment Key Laboratory of Higher Education Institutions in Shaanxi Province, School of Life Science and Technology, Xidian University, Xi'an, Shaanxi 710126, China (e-mail: xlchen@xidian.edu.cn).}
\thanks{S. Kevin Zhou is with School of Biomedical Engineering \& Suzhou Institute for Advanced Research, Center for Medical Imaging, Robotics, Analytic Computing \& Learning (MIRACLE), University of Science and Technology of China, Suzhou, China (e-mail: skevinzhou@ustc.edu.cn).}
\thanks{Wuwei Ren, Jingyi Yu, and Yuyao Zhang (\textit{corresponding author}) are with School of Information Science and Technology and Shanghai Engineering Research Center of Intelligent Vision and Imaging, ShanghaiTech University, Shanghai 201210, China (e-mail: \{renww, yujingyi, zhangyy8\}@shanghaitech.edu.cn).}}

\maketitle
\begin{abstract}
\par X-ray CT often suffers from shadowing and streaking artifacts in the presence of metallic materials, which severely degrade imaging quality. Physically, the linear attenuation coefficients (LACs) of metals vary significantly with X-ray energy, causing a nonlinear beam hardening effect (BHE) in CT measurements. Reconstructing CT images from metal-corrupted measurements consequently becomes a challenging nonlinear inverse problem. Existing state-of-the-art (SOTA) metal artifact reduction (MAR) algorithms rely on supervised learning with numerous paired CT samples. While promising, these supervised methods often assume that the unknown LACs are energy-independent, ignoring the energy-induced BHE, which results in limited generalization. Moreover, the requirement for large datasets also limits their applications in real-world scenarios. In this work, we propose Density neural representation (Diner), a novel unsupervised MAR method. Our key innovation lies in formulating MAR as an energy-independent density reconstruction problem that strictly adheres to the photon-tissue absorption physical model. This model is inherently nonlinear and complex, making it a rarely considered approach in inverse imaging problems. By introducing the water-equivalent tissues approximation and a new polychromatic model to characterize the nonlinear CT acquisition process, we directly learn the neural representation of the density map from raw measurements without using external training data. This energy-independent density reconstruction framework fundamentally resolves the nonlinear BHE, enabling superior MAR performance across a wide range of scanning scenarios. Extensive experiments on both simulated and real-world datasets demonstrate the superiority of our unsupervised Diner over popular supervised methods in terms of MAR performance and robustness. To the best of our knowledge, Diner is the first unsupervised MAR method to outperform its supervised counterparts.
\end{abstract}
\begin{IEEEkeywords}
X-ray CT, Metal Artifact Reduction, Neural Representation, NeRF, Unsupervised Learning
\end{IEEEkeywords}
\section{Introduction}
\par \IEEEPARstart{X}{-ray} computed tomography (CT) is a vital biomedical imaging technique for visualizing the anatomy of objects. CT imaging aims to determine the spatial distribution of the linear attenuation coefficients (LACs) of objects to X-rays. While the LACs of biological tissues remain nearly constant across X-ray energies, those of metals exhibit significant variations. This physical discrepancy introduces a substantial nonlinear beam hardening effect (BHE) in CT measurements when metals are present~\cite{seo2012nonlinear}. Consequently, recovering the LAC map from metal-affected measurements becomes a complex nonlinear inverse problem. Traditional linear algorithms, such as FBP~\cite{fbp}, often produce severe shadowing and streaking artifacts in CT images. These artifacts can significantly compromise the reliability of clinical diagnoses and biomedical research, including applications such as bone analysis.
\par Metal artifact reduction (MAR) has long been a critical and enduring challenge in X-ray CT~\cite{gjesteby2016metal}. Model-based methods~\cite{kalender1987li, art_method} approach MAR by formulating it as an image inpainting problem. These methods treat metal-corrupted signals in the measurements as missing data and employ various schemes (\eg linear interpolation) to fill in the gaps. However, such frameworks often introduce noticeable secondary artifacts in the reconstructed images, as the interpolation algorithms fail to satisfy the geometric constraints of CT scanning. Recently, supervised deep learning (DL) models~\cite{cnnmar, lin2019dudonet, wang2022acdnet, wang2021dicdnet, adn2019_tmi, yu2020deep, wang2018conditional} have dominated the MAR problem. In these methods, deep neural networks are trained to map metal-affected measurements to clean measurements or artifact-corrupted images to artifact-free images. By leveraging the capabilities of neural networks and data-driven priors, supervised DL methods achieve state-of-the-art (SOTA) MAR performance. However, they face two major limitations: 1) Supervised learning requires a large-scale training dataset with numerous paired samples, which is extremely challenging to collect; 2) These methods are prone to performance degradation when CT acquisition protocols (\eg variations in metal shapes and X-ray acquisition geometry) deviate from those used in the training dataset, leading to the out-of-domain (OOD) problem. These limitations significantly hinder the practical application of supervised MAR models in real-world scenarios.
\par Implicit neural representation (INR) has recently emerged as a revolutionary unsupervised DL framework for various CT reconstruction tasks, such as sparse-view CT~\cite{shen2022nerp, wu2022self, sun2021coil, wu2022joint, zha2022naf, ruckert2022neat}, limited-angle CT~\cite{zang2021intratomo, ruckert2022neat, du2024dper}, and dynamic CT~\cite{chen2022motion, gupta2023difir, reed2021dynamic}. INR represents an unknown CT image as a continuous function of spatial positions. By incorporating a differentiable linear integral model (\ie Radon transform), it optimizes a multi-layer perceptron (MLP) to approximate the function by minimizing the errors in the measurements. Due to the inherent learning bias of MLPs toward low-frequency image patterns~\cite{rahaman2019spectral, xu2019frequency}, INR can generate high-quality CT images under highly ill-posed conditions. Recently, Wu et al.\cite{wu2023unsupervised} proposed Polyner, an INR-based MAR method. Its core innovation is the introduction of a polychromatic physical model to characterize the nonlinear CT acquisition process. By using the polychromatic forward model, Polyner\cite{wu2023unsupervised} optimizes INR to reconstruct polychromatic LACs at multiple X-ray energies from raw measurements. This approach significantly mitigates the energy-induced BHE and achieves promising results. However, predicting multiple LAC maps poses a highly ill-posed inverse problem. To address this, Polyner~\cite{wu2023unsupervised} requires a local energy smoothness regularizer to constrain the solution space, which significantly limits its MAR performance and robustness.
\par In this work, we propose \textbf{D}ensity \textbf{ne}ural \textbf{r}epresentation (\textbf{Diner}), a new unsupervised method that formulates MAR as an energy-independent density reconstruction problem. This formulation fundamentally addresses the nonlinear BHE, resulting in superior MAR reconstructions. Conceptually, we decompose energy-dependent LACs into the product of energy-dependent mass attenuation coefficients (MACs) and energy-independent densities using a standard physical model~\cite{mcnair1981icru, beutel2000handbook}. By further employing the water-equivalent tissue approximation~\cite{huang1976evaluation, Geraldelli} (\ie the MACs of biological tissues are approximately equal to those of water), we can leverage prior knowledge of the known MACs of water and metal. Therefore, our objective is to directly reconstruct densities from metal-affected measurements. Compared to Polyner~\cite{wu2023unsupervised}, which estimates multiple polychromatic LAC maps, our Diner substantially reduces the ill-posed nature of the inverse problem by reconstructing a single energy-independent density map, achieving enhanced reconstructions.
\par To achieve stable optimization, we adopt INR as the core framework. Specifically, we represent the unknown density map as an implicit function of spatial coordinates. Subsequently, we introduce a novel differentiable polychromatic forward model that converts the densities into CT measurements while enabling gradient back-propagation. This allows us to optimize an MLP network to learn the function by minimizing the error between the estimated and actual measurements. Our Diner is a fully unsupervised technique, eliminating the need for external training data and thus enhancing its potential for diverse clinical applications. We evaluate Diner on four datasets, including two simulated and two real-world datasets. Empirical results demonstrate that our unsupervised Diner slightly outperforms well-known supervised MAR techniques on in-domain datasets and significantly surpasses them on out-of-domain datasets. The main contributions of this work are as follows:
\begin{itemize}
    \item We propose Diner, an unsupervised, physical-model-driven MAR approach capable of recovering high-quality CT images from metal-corrupted measurements without requiring external training data.
    \item We introduce a novel formulation for the nonlinear MAR problem by solving the energy-independent density map, which fundamentally addresses the nonlinear, energy-induced BHE.
    \item We present a differentiable physical forward model that accurately simulates the nonlinear physical acquisition process from densities to CT measurements.
    \item We integrate the proposed physical forward model into the INR framework, enabling it to effectively address nonlinear MAR problems.
\end{itemize}

\section{Preliminaries}
\par The Lambert-Beer's law~\cite{lambert1760photometria, beer1852bestimmung} formulates the physical process of an X-ray passing through an object as below:
\begin{equation}
    I(\mathbf{r}, E) = I_0(E)\cdot\text{exp}\left(-\int_\mathbf{r}\mu(\mathbf{x}, E)\mathrm{d}\mathbf{x}\right),
\end{equation}
where $\mathbf{r}$ represents a monochromatic (\ie single-energy) X-ray, $E$ is its energy level, $I_0(E)$ and $I(\mathbf{r}, E)$ respectively are the numbers of photons emitted by an X-ray source and received by an X-ray detector, and $\mu(\mathbf{x}, E)$ is the LAC of the object at the position $\mathbf{x}$. The LAC value serves as a measure of the capacity of the object to absorb the X-ray.
\par The CT measurements $\rho(\mathbf{r})$ are typically defined as a linear integral of the LACs $\mu$ along the X-ray $\mathbf{r}$:
\begin{equation}
    \rho(\mathbf{r})\triangleq-\ln\frac{I(\mathbf{r}, E)}{I_0(E)}=\int_\mathbf{r}\mu(\mathbf{x}, E)\mathrm{d}\mathbf{x}.
    \label{eq. measurement define}
\end{equation}
The CT reconstruction problem is to solve the spatial distribution of the LACs $\mu(\mathbf{x}, E)$ from the measurement data $\rho(\mathbf{r})$. When the Nyquist-Shannon sampling theorem is satisfied~\cite{1697831}, standard techniques (\eg FBP~\cite{fbp}) for linear inverse problems can produce high-quality CT images.
\par However, most clinical CT scanners emit polychromatic X-rays (\ie within a range of energy levels) due to physical limitations of the X-ray source hardware~\cite{seo2012nonlinear}. Given an X-ray composed of multiple monochromatic X-rays within a range of $\mathcal{E}$, the CT measurement data can be expressed by a nonlinear polychromatic model:
\begin{equation}
\begin{aligned}
       \rho(\mathbf{r}) &=-\ln\frac{\int_\mathbf{\mathcal{E}}I(\mathbf{r}, E)\mathrm{d}E}{\int_\mathbf{\mathcal{E}}I_0(E)\mathrm{d}E}\\
       &=-\ln\int_\mathcal{E}\eta(E)\cdot\mathrm{exp}\left(-\int_{\mathbf{r}}\mu(\mathbf{x}, E)\mathrm{d}\mathbf{x}\right)\mathrm{d}E,
\end{aligned}
   \label{eq:ct-forward}
\end{equation}
where $\eta(E)={I_0(E)}/{\int_\mathcal{E}I_0(E')\mathrm{d}E'}$ represents the normalized energy spectrum that characterizes the distribution of the number of photons $I_0(E)$ emitted by the X-ray source over the energy level $E$.
\par The LACs of the biological tissue almost remain constant with increasing the energy level of the X-ray~\cite{cnnmar, lin2019dudonet, adn2019_tmi}. A common assumption for the biological tissue is:
\begin{equation}
    |\mu(\mathbf{x}, E_a)-\mu(\mathbf{x}, E_b)|\approx0,\ \forall E_a,E_b\in\mathcal{E},
    \label{eq. body assumption}
\end{equation}
\ie the LACs of the biological tissue are considered energy-independent. Thus, we can represent their LACs as a function of spatial coordinates $\mu(\mathbf{x})$ and then derive the corresponding measurement data as below:
\begin{equation}
\begin{aligned}
       \rho(\mathbf{r}) =-\ln\int_\mathcal{E}\eta(E)\cdot\mathrm{exp}\left(-\int_{\mathbf{r}}\mu(\mathbf{x})\mathrm{d}\mathbf{x}\right)\mathrm{d}E 
       = \int_{\mathbf{r}}\mu(\mathbf{x})\mathrm{d}\mathbf{x}.
\end{aligned}
\end{equation}
This implies that the CT acquisition for the biological tissue can still be approximated as a linear integral transformation when even using polychromatic X-rays. Therefore, the CT images by FBP~\cite{fbp} are artifacts-free in this case.
\par However, the LACs of metallic materials undergo sharp variations as a function of the X-ray energies~\cite{seo2012nonlinear}. For an organism with metallic implants, we thus have to split its LAC map into two parts: tissue region and metal region. Mathematically, it can be expressed as: 
\begin{equation}
    \mu(\mathbf{x}, E)=\left[1-\mathcal{M}(\mathbf{x})\right]\cdot\mu_\mathrm{tissue}(\mathbf{x})+\mathcal{M}(\mathbf{x})\cdot\mu_\mathrm{metal}(\mathbf{x}, E),
    \label{eq: mask lacs}
\end{equation}
where $\mu_\mathrm{tissue}(\mathbf{x})$ and $\mu_\mathrm{metal}(\mathbf{x}, E)$  denote the LACs of the tissue and metal, respectively. $\mathcal{M}$ is a metal binary mask ($\mathcal{M}=1$ for the metal region and $\mathcal{M}=0$ otherwise). 
\par Substituting Eq.~\eqref{eq: mask lacs} into Eq.~\eqref{eq:ct-forward}, the CT measurement data can be derived as below:
\begin{equation}
\begin{aligned}
    \rho(\mathbf{r}) =&\overbrace{\int_{\mathbf{r}}\left[1-\mathcal{M}(\mathbf{x})\right]\cdot\mu_\mathrm{tissue}(\mathbf{x})\mathrm{d}\mathbf{x}}^{\text{\small{Linear Integral}}}\\
    &-\underbrace{\ln\int_\mathcal{E}\eta(E)\cdot\mathrm{exp}\left(-\int_{\mathbf{r}}\mathcal{M}(\mathbf{x})\cdot\mu_\mathrm{metal}(\mathbf{x}, E)\mathrm{d}\mathbf{x}\right)\mathrm{d}E}_{\text{\small{Nonlinear Beam Hardening Effect}}}.
\end{aligned}
    \label{eq:nonlinear}
\end{equation}
Eq.~\eqref{eq:nonlinear} means that the CT measurements of an organism containing metallic materials should be divided into two parts: (1) a linear integral for the tissue and (2) an energy-induced non-linear transform for the metals. The latter results in a nonlinear beam hardening effect (BHE) in the CT measurements. In this case, the use of traditional linear FBP~\cite{fbp}, will result in severe metal artifacts in CT images.
\begin{figure}[t]
    \centering
    \includegraphics[width=0.4\textwidth]{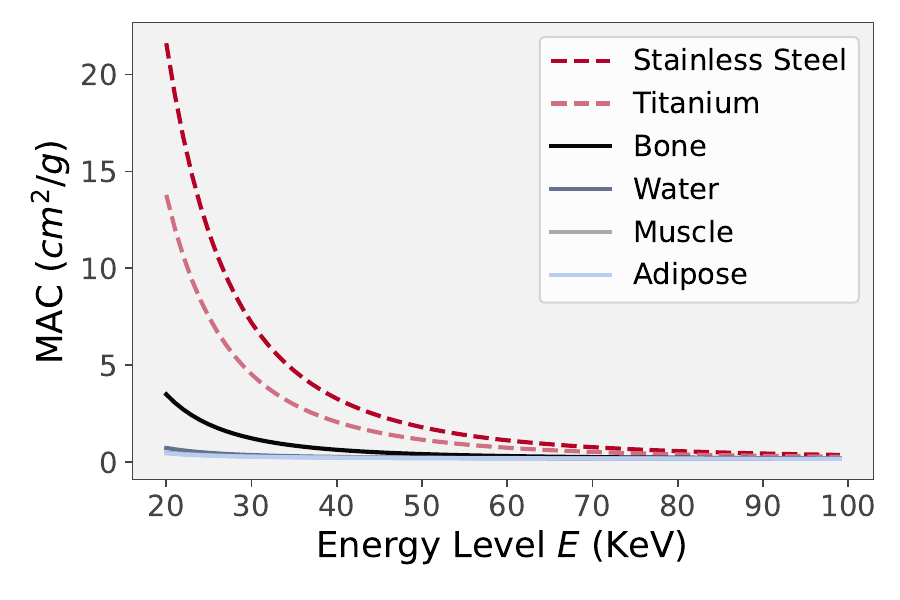}
    \caption{MAC curves of different materials over the X-ray energy range $E$ of [20, 100] KeV. The data is from the XCOM database~\cite{berger2009xcom}.}
    \label{fig:mac_curve}
\end{figure}
\section{Methodology}
\subsection{Problem Formulation}
\par As illustrated in Eq.~\ref{eq:nonlinear}, the nonlinear BHE is the key factor for addressing the CT MAR problem. Previous MAR approaches~\cite{rudin1992nonlinear, cnnmar, song2022solving, wang2022acdnet,wang2021dicdnet, kalender1987li} mostly assume that the LACs to be resolved are energy-independent and reconstruct them at a virtual monochromatic X-ray energy level. However, this framework fails to capture the energy-induced BHE, which results in suboptimal MAR performance.
\par In this work, we propose directly reconstructing the densities $\sigma$ of observed objects. Since the densities $\sigma$ are physically energy-independent, this approaches can fundamentally address the energy-induced BHE and significantly improve MAR performance. However, as shown in Eq.~\ref{eq:ct-forward}, the CT measurement data $\rho$ is a nonlinear transform of the energy-dependent LACs $\mu$, and is not directly related to the densities $\sigma$. To establish an explicit relationship between the measurement data $\rho$ and densities $\sigma$, we introduce a standard physical model~\cite{mcnair1981icru, beutel2000handbook} to decompose the LACs $\mu$ as a product of the energy-independent densities $\sigma$ and energy-dependent mass attenuation coefficients (MACs) $\gamma$, as below:
\begin{equation}
    \mu(\mathbf{x}, E) = \gamma(\mathbf{x}, E)\cdot\sigma(\mathbf{x}).
    \label{eq: mac-lac}
\end{equation}
\par For a organism containing metallic implants, we leverage two basic and reasonable assumptions:
\begin{assumption}
\label{ass:1}
(Water-Equivalent Tissues Approximation~\cite{huang1976evaluation, Geraldelli}) The MACs of biological tissues are approximately replaced by those of water, \ie $\gamma(\mathbf{x}, E) \triangleq \gamma_\mathrm{water}(E),\forall \mathbf{x}\notin \mathcal{M}$, where $\mathcal{M}$ denotes metal regions.
\end{assumption}
\begin{assumption}
\label{ass:2}
The metallic implants are homogeneous such that $|\gamma(\mathbf{x}_a, E)-\gamma(\mathbf{x}_b, E)|\approx0,\forall \mathbf{x}_a,\mathbf{x}_b\in\mathcal{M}$ holds, and their position-independent MACs are accessible.
\end{assumption}
\begin{remark}
\par Experimental findings~\cite{huang1976evaluation, Geraldelli} have shown that most types of biological tissue (\eg adipose tissue and muscle tissue) have similar MACs that are very close to those of water. Figure~\ref{fig:mac_curve} shows the MAC curves of two metals (titanium and 304 stainless steel), three types of human tissue (muscle, bone, and adipose), and water over the X-ray energy range of [20, 100] KeV. We observe that the three human tissues are significantly closer to water than the two metals. 
\end{remark}
\begin{remark}
Most common biological and clinical metallic implants (\eg titanium and 304 stainless steel) are homogeneous, and their corresponding MACs can be easily accessible in some well-known open-source databases, such as NIST standard library~\cite{HubbellSeltzer2004} and XCOM program~\cite{berger2009xcom}.
\end{remark}
\par According to assumptions~\ref{ass:1} and~\ref{ass:2}, by substituting Eq.~\ref{eq: mac-lac} into Eq.~\ref{eq: mask lacs}, we define a novel LAC decomposition model for the CT MAR problem as below:
\begin{equation}
    \begin{aligned}
        \mu(\mathbf{x}, E) = \sigma(\mathbf{x})\cdot \big\{[1-&\mathcal{M}(\mathbf{x})]\cdot\gamma_\mathrm{water}(E)+\\
        &\mathcal{M}(\mathbf{x})\cdot\gamma_\mathrm{metal}(E)\big\},
    \end{aligned}
    \label{eq: lacs-density}
\end{equation}
where $\gamma_\mathrm{water}$ and $\gamma_\mathrm{metal}$ respectively denote the known MACs of the water and metals. Eq.~\ref{eq: lacs-density} implies that the energy-dependent LACs $\mu(\mathbf{x}, E)$ can be reconstructed accurately by estimating the energy-independent densities $\sigma(\mathbf{x})$. Therefore, the CT MAR is formulated as an energy-independent density reconstruction problem (\ie $\rho\rightarrow\sigma$).
\begin{figure*}
    \centering
    \includegraphics[width=0.95\textwidth]{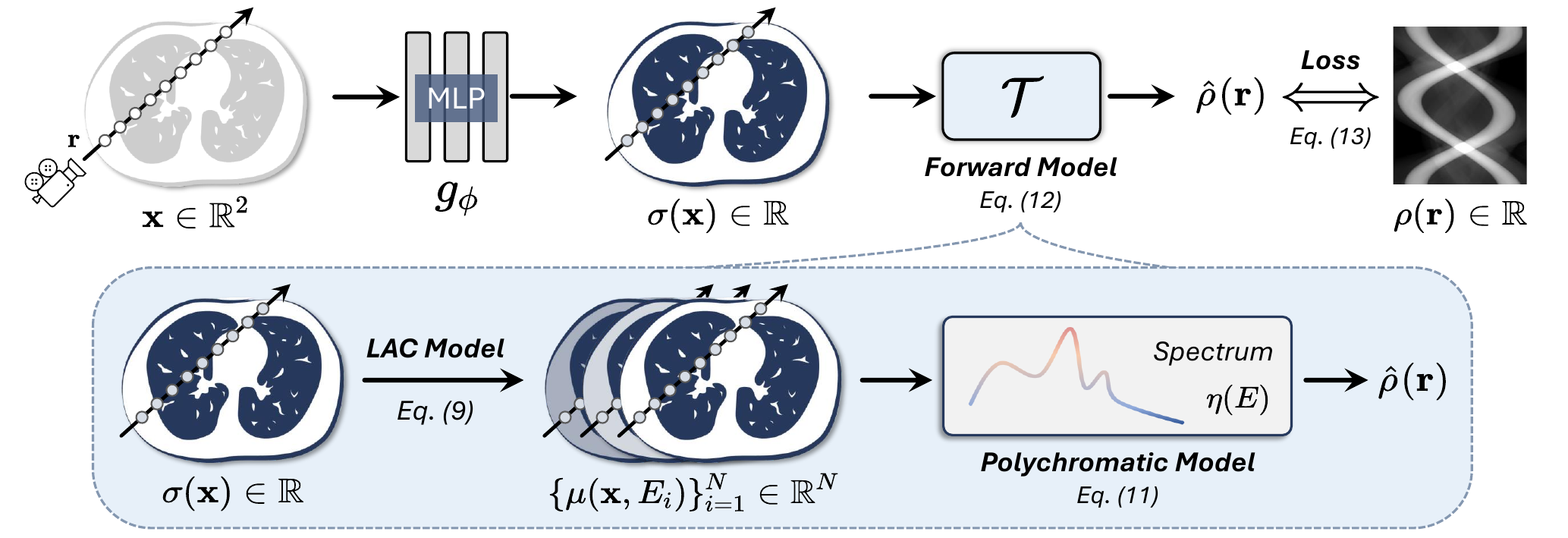}
    \caption{Overview of the proposed Diner model. For an X-ray $\mathbf{r}$, we first sample a set of coordinates $\mathbf{x}$ at a fixed interval $\Delta\mathbf{x}$. Then, we feed these coordinates $\mathbf{x}$ into an MLP network $g_\phi$ to predict the densities $\sigma(\mathbf{x})$ of the observed object at these positions. Furthermore, we use the proposed nonlinear forward model $\mathcal{T}$ (Eq.~\ref{eq: full-forward-model}) to transform these MLP-predicted densities $\sigma(\mathbf{x}), \forall\mathbf{x}\in\mathbf{r}$ into measurement data $\hat{\rho}(\mathbf{r})$. Specifically, these densities $\sigma(\mathbf{x})$ are first converted into polychromatic LACs $\{\mu(\mathbf{x}, E_i)\}_{i=1}^N$ using our LAC decomposition model (Eq.~\ref{eq: lacs-density}). Then, the measurement $\hat{p}(\mathbf{x})$ is generated from these LACs using the polychromatic model (Eq.~\ref{eq: dis-forward-model}). Finally, we minimize the predicted errors $\mathcal{L}_\text{DC}$ between the estimated $\hat{\rho}(\mathbf{r})$ and raw $\rho(\mathbf{r})$ measurements to optimize the trainable weights of the MLP network $g_\phi$ for learning the density neural representation.}
    \label{fig:method}
\end{figure*}
\subsection{Solving Density via Neural Representation}
\par Figure~\ref{fig:method} shows the workflow of solving the densities via the proposed Diner model. In this section, we introduce the main technical contributions of our Diner in detail.
\subsubsection{\textbf{MLP-based Parameterization}}
\par Without loss of generality, we introduce Diner model based on 2D CT acquisition systems for simplicity here. However, the proposed technique can be easily extended to more advanced CT protocols (\eg 3D cone-beam and helical CT) through a flexible geometry implementation. Given a 2D CT acquisition system, we represent the density map to be reconstructed as a function of spatial coordinates, defined by
\begin{equation}
    f: \mathbf{x}=(x,y)\in\mathbb{R}^2 \rightarrow \sigma\in\mathbb{R},
\end{equation}
where $\mathbf{x}$ is any spatial coordinate and $\sigma$ is the density of the object at the position $\mathbf{x}$. However, solving the analytical expression of the implicit function $f$ is intractable. 
\par Inspired by the recent advancements of the INR framework~\cite{sitzmann2020implicit, mildenhall2021nerf, tancik2020fourier, muller2022instant} for recovering signals, we employ an MLP network $g_\phi$ that takes a single coordinate $\mathbf{x}$ and outputs the corresponding density $\sigma$ to approximate the function (\ie $f\approx g_\phi$). Technically, we optimize the trainable weights $\phi$ of the MLP network. Once the optimization is completed, the desired density map is parameterized into the weights $\phi$. Taking advantage of the MLP network's powerful fitting capability and learning bias for low-frequency signals~\cite{rahaman2019spectral,xu2019frequency}, such MLP-based parameterization can yield high-quality reconstructions in line with natural image manifolds.
\subsubsection{\textbf{Differentiable Forward Model}}
\par The output of the MLP network $g_\phi$ is expected as the density $\sigma$, a physical property of the observed object. We thus need to define a new physical forward model $\mathcal{T}$ to simulate the complex acquisition process from the underlying property (\ie density) to the raw CT measurements. Our forward model includes two stages: 
\begin{itemize}
    \item Performing our LAC decomposition model (Eq.~\ref{eq: lacs-density}) to transform the MLP-predicted, energy-independent densities $\sigma(\mathbf{x}) = g_{\phi}(\mathbf{x})$ into the polychromatic LACs $\mu(\mathbf{x}, E)$ at multiple energy levels;
    \item Using the discrete version of the polychromatic model (Eq.~\ref{eq:ct-forward}) to generate the estimated measurement $\hat{\rho}(\mathbf{r})$ from the polychromatic LACs $\mu(\mathbf{x}, E)$.
\end{itemize}
Formally, the discrete polychromatic model is defined as:
\begin{equation}
    \hat{\rho}(\mathbf{r}) = -\ln\sum_{i=1}^{N}\eta(E_i)\cdot\text{exp}\left(-\sum_{\mathbf{x}\in\mathbf{r}}\mu(\mathbf{x}, E_i)\cdot\Delta\mathbf{x}\right),
    \label{eq: dis-forward-model}
\end{equation}
where $\Delta\mathbf{x}$ denotes the Euclidean distance between adjacent sampling coordinates along the X-ray $\mathbf{r}$. The normalized energy spectrum $\eta\in\mathbb{R}^N$ is considered a known prior knowledge. We use the SPEKTR toolkit~\cite{punnoose2016technical} to estimate it. Its effects on the MAR reconstruction performance are explored in the following experiments.
\par In summary, our full forward model $\mathcal{T}$ is defined by:
\begin{equation}
    \mathcal{T}:\ \sigma(\mathbf{x})\xrightarrow{\text{Eq.~\ref{eq: lacs-density}}} \mu(\mathbf{x}, E)\xrightarrow{\text{Eq.~\ref{eq: dis-forward-model}}} \hat{\rho}(\mathbf{r}).
    \label{eq: full-forward-model}
\end{equation}
It is worth noting that this forward model $\mathcal{T}$ is differentiable, enabling the use of gradient backpropagation algorithms to optimize the MLP network $g_\phi$.
\subsubsection{\textbf{Loss Function}}
\par By leveraging the forward model $\mathcal{T}$, the MLP-predicted densities $\sigma(\mathbf{x}) = g_{\phi}(\mathbf{x}),\forall \mathbf{x}\in\mathbf{r}$ are transformed into the CT measurement data $\hat{\rho}(\mathbf{r})$. Then, we define a loss function $\mathcal{L}_\text{DC}$ to compute the mean absolute error between the estimated $\hat{\rho}$ and real measurements ${\rho}$. The loss function can provide data consistency with the real measurements $\rho$. It can be expressed as below: 
\begin{equation}
\mathcal{L}_{\text{DC}} = \frac{1}{\left |\mathcal{R} \right |}\sum_{\mathbf{r}\in\mathcal{R}} \left | \hat{\rho}(\mathbf{r})- \rho(\mathbf{r})\right |,
\label{eq: loss-function}
\end{equation}
where $\mathcal{R}$ denotes a set of randomly sampling X-rays $\mathbf{r}$ at each training iteration.
\subsection{Virtual Monochromatic CT Image Reconstruction}
\par After the optimization, we feed all spatial coordinates $\mathbf{x}$ into the well-trained MLP network to produce the corresponding densities $\sigma(\mathbf{x})=g_{\phi^*}(\mathbf{x})$, meaning a density map $\sigma(\cdot)$ can be reconstructed. Then, we use our LACs decomposition model (Eq.~\ref{eq: lacs-density}) to generate the energy-dependent LAC maps $\mu(\cdot, E)$ at $N$ energy levels (\ie polychromatic CT images). However, this work aims to recover a single monochromatic artifacts-free CT image. Consider a polychromatic X-ray composed of multiple monochromatic X-rays at $N$ energy levels, the energy level of its equivalent monochromatic X-ray is given by 
\begin{equation}
    E^\star=\left \lfloor\sum_{i=1}^{N}\eta(E_i)\cdot E_i\right \rfloor,
\end{equation}
where $\left \lfloor \cdot \right \rfloor$ represents floor function. Finally, the LAC map $\mu(\cdot, E^\star)$ at the energy level $E^\star$ is employed for the artifact-free reconstruction output.

\subsection{Implementation Details}
\par For our Diner, we employ the recent hash encoding~\cite{muller2022instant} followed by two fully connected (FC) layers to implement the MLP network $g_\phi$. The hash encoding can effectively accelerate the training process. Its hyper-parameters are set as below: $L=16$, $T=2^{19}$, $F=8$, $N_\mathrm{min}=2$ and $b=2$. The two FC layers have 128 neurons, followed by ReLU and ELU activation functions, respectively. At each training iteration, we set the number of the random sampling X-rays as 80, \ie $|\mathcal{R}|=80$ in Eq.~\ref{eq: loss-function}. We use the Adam optimizer with default hyperparameters. The learning rate starts from 1e-3 and decays by half per 500 epochs. The total training epochs is 2000, which takes about 1.2 minutes on a single NVIDIA RTX TITIAN GPU. Note that all hyper-parameters are tuned on 10 samples from the DeepLesion dataset~\cite{deeplesion} and then fixed across all other samples.

\begin{table}[t]
    \setlength{\tabcolsep}{1mm}
    \caption{Detailed parameters of the acquisition processes for the DeepLesion~\cite{deeplesion} and LIDC~\cite{LIDC} simulation datasets and mouse tight and body phantom real-world datasets.}
    \centering
    \begin{tabular}{cccc}
    \toprule
      \multirow{2.5}{*}{\textbf{Parameters}} & \multicolumn{1}{c}{\textbf{Simulation Data}} & \multicolumn{2}{c}{\textbf{Real-world Data}}\\ \cmidrule{2-4}
         & DeepLesion\texttt{/}LIDC &  Mouse Tight & Body Phantom\\ \midrule
        Manufacturer & -- & \textsc{Bruker} & \textsc{UIH} \\
        Geometry & 2D Fan-beam & 3D Cone-beam& 3D Helical\\
        Image Size & 256$\times$256 & 500$\times$500$\times$500 & 512$\times$512$\times$490 \\
        Voxel Size & 1$\times$1 mm$^2$ & 0.03$^3$ mm$^3$ & 0.5$^3$ mm$^3$\\
        Source Voltage & -- & 60 kV & 120 kV\\
        Scanning Range & [0, 360)$^\circ$ & [0, 360)$^\circ$ & [0, 360]$^\circ$\\
        Angular Spacing & 0.1$^\circ$ & --&  --\\
        Detector Spacing & -- & 0.07 mm & 0.5 mm\\
        Source to Center & 362 mm & 92.60 mm& 570 mm\\
        Center to Detector & 362 mm & 65.95 mm& 490 mm\\
        Number of Angles & 360\texttt{/}270 & 900& 48,867\\
     \bottomrule
    \end{tabular}
    \label{tab:geometry}
\end{table}


\section{Experiments}
\par We conduct comprehensive experiments to assess the effectiveness of our proposed approach, including \romannumeral1) Comparison with the SOTA MAR methods on both simulated and real-world datasets, \romannumeral2) Extensive ablation studies to analyze and understand the components and intricacies of our Diner, and \romannumeral3) An examination of failure cases and limitations of our Diner. The empirical studies confirm the superiority of our unsupervised Diner over existing MAR methods in terms of MAR performance and robustness.
\subsection{Dataset and Pre-processing}
\par We conduct experiments on four diverse datasets, including two simulation datasets and two real-world datasets. Table~\ref{tab:geometry} shows the detailed parameters of the four datasets.
\subsubsection{\textbf{DeepLesion and LIDC Simulation Datasets}}
\par DeeLesion~\cite{deeplesion} and LIDC~\cite{LIDC} are the two most commonly used datasets for the CT MAR evaluation. For the two datasets, we extract 50 and 20 two-dimensional (2D) slices, respectively, from raw three-dimensional (3D) CT volumes as ground truth (GT) samples. Then, we follow the data pre-processing pipeline presented in the prior studies~\cite{adn2019_tmi, cnnmar, lin2019dudonet} to synthesize metal-corrupted measurements as input data. Specifically, we use ten shapes of metals provided by Zhang \etal~\cite{cnnmar} and simulate three types of common medical materials (including titanium, chromium, and 304 stainless steel) according to the XCOM database~\cite{berger2009xcom}. We simulate a polychromatic X-ray source by following literature~\cite{cnnmar, adn2019_tmi, wang2022acdnet, wang2021dicdnet}. We also incorporate Poisson noise and the partial volume effect into the measurement data. Table~\ref{tab:geometry} shows the detailed parameters of the acquisition processes for the two simulation datasets. It is worth noting that all data are solely used for the test since our method is fully unsupervised.
\begin{figure}[t]
    \centering
    \includegraphics[width=0.475\textwidth]{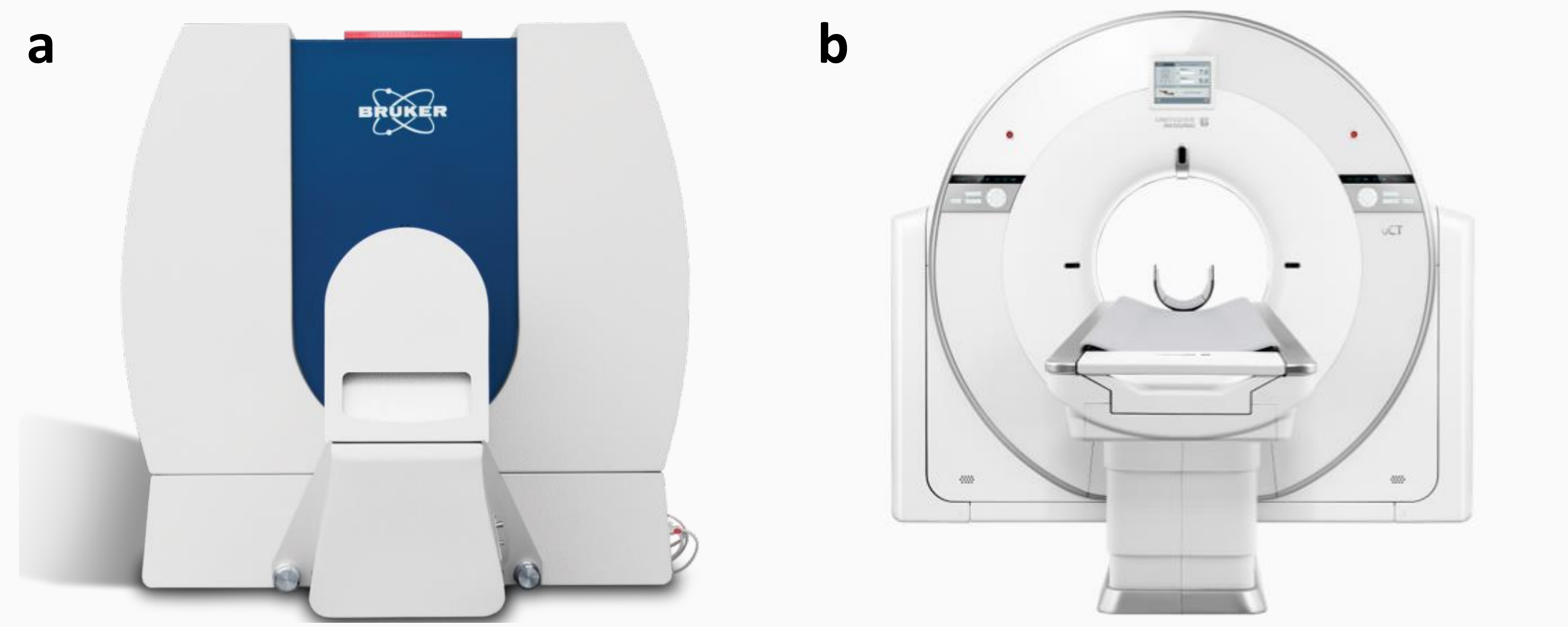}
    \caption{\textbf{a)} A commercial Bruker SKYSCAN 1276 micro-CT scanner used in our mouse tight data acquisition and \textbf{b)} A commercial UIH uCT 768 scanner used in our body phantom data acquisition.}
    \label{fig:fig_microCT}
\end{figure}
\begin{table*}[t]
\centering
\caption{Quantitative results of compared methods for three types of metals on the DeepLesion~\cite{deeplesion} and LIDC~\cite{LIDC} simulation datasets. The best and second performances are respectively highlighted in \textbf{bold} and \underline{underline}. The subscripts $\spadesuit$, $\heartsuit$, $\diamondsuit$, and $\clubsuit$ respectively represent model-based, unpaired-supervised, supervised, and unsupervised MAR techniques.}
\label{table_comparison}
\begin{tabular}{clcccccc} 
\toprule 
\multirow{4}{*}{\textbf{Dataset}}     & \multirow{4}{*}{\textbf{Method}} & \multicolumn{6}{c}{\textbf{Types of Metals}} \\
\cmidrule{3-8}
                              &                         & \multicolumn{2}{c}{\textsc{Titanium}}   & \multicolumn{2}{c}{\textsc{Chromium}} & \multicolumn{2}{c}{\textsc{304 Stainless Steel}}\\ \cmidrule(r){3-4}\cmidrule(r){5-6}\cmidrule(r){7-8}
&                         & PSNR           & SSIM            & PSNR           & SSIM             & PSNR           & SSIM\\ 
\midrule
\multirow{11}{*}{\texttt{DeepLesion}~\cite{deeplesion}}  & FBP$^\spadesuit$~\cite{fbp}                     & 30.61$\pm$2.71               & 0.7829$\pm$0.0804                & 26.67$\pm$2.89               & 0.6465$\pm$0.1052  & 25.01$\pm$2.95    & 0.5888$\pm$0.1098\\
& LI$^\spadesuit$~\cite{kalender1987li}                     & 32.07$\pm$2.61              & 0.8650$\pm$0.0518                & 32.07$\pm$2.61               & 0.8650$\pm$0.0518 & 32.07$\pm$2.61 & 0.8650$\pm$0.0518  \\
& ART$^\spadesuit$~\cite{art_method}                     & 32.60$\pm$3.02              & 0.8502$\pm$0.0554                & 32.60$\pm$3.02              & 0.8502$\pm$0.0554 & 32.60$\pm$3.02              & 0.8502$\pm$0.0554  \\
\cmidrule{2-8}
& ADN$^\heartsuit$~\cite{adn2019_tmi}            & 33.83$\pm$2.48               & 0.9508$\pm$0.0156                & 32.58$\pm$3.16              & 0.9409$\pm$0.0223   & 31.77$\pm$3.54  & 0.9340$\pm$0.0263\\
\cmidrule{2-8}
   & CNN-MAR$^\diamondsuit$~\cite{cnnmar}                & 35.05$\pm$2.31               & 0.9385$\pm$0.0218                & 34.71$\pm$2.38              & 0.9322$\pm$0.0257       & 34.50$\pm$2.44        & 0.9277$\pm$0.0289 \\
                              & DICDNet$^\diamondsuit$~\cite{wang2021dicdnet}                & 37.93$\pm$1.89               & 0.9710$\pm$0.0094                & 37.18$\pm$2.21               & 0.9654$\pm$0.0128    & 36.74$\pm$2.47 & 0.9623$\pm$0.0148 \\
                              & ACDNet$^\diamondsuit$~\cite{wang2022acdnet}                 & \underline{38.43$\pm$1.91}               & 0.9701$\pm$0.0112               & \underline{37.46$\pm$2.25}              & 0.9586$\pm$0.0171   & \underline{36.90$\pm$2.50} & 0.9524$\pm$0.0200           \\ 
\cmidrule{2-8}
& Polyner$^\clubsuit$~\cite{wu2023unsupervised}                    & 37.23$\pm$1.62             & \underline{0.9769$\pm$0.0048}                 & 36.96$\pm$2.00               & \underline{0.9754$\pm$0.0057}  & 36.65$\pm$2.48 & \underline{0.9739$\pm$0.0069}\\
& Diner$^\clubsuit$ (Ours)                    & \textbf{39.17$\pm$1.85}               & \textbf{0.9846$\pm$0.0035}               & \textbf{38.77$\pm$1.81}               & \textbf{0.9824$\pm$0.0043}   & \textbf{38.57$\pm$1.78}
     & \textbf{0.9814$\pm$0.0048}\\ \midrule
\multirow{11}{*}{\texttt{LIDC}~\cite{LIDC}}  & FBP$^\spadesuit$~\cite{fbp}                     & 25.87$\pm$2.23               & 0.5435$\pm$0.1219                & 22.20$\pm$2.43             & 0.4063$\pm$0.1201  & 20.63$\pm$2.54    & 0.3573$\pm$0.1158\\
& LI$^\spadesuit$~\cite{kalender1987li}                     & 29.30$\pm$2.38              & 0.7457$\pm$0.0941                & 29.30$\pm$2.38              & 0.7457$\pm$0.0941 & 29.30$\pm$2.38              & 0.7457$\pm$0.0941  \\
& ART$^\spadesuit$~\cite{art_method}                     & 27.99$\pm$2.97              & 0.6565$\pm$0.1069                & 27.99$\pm$2.97              & 0.6565$\pm$0.1069 & 27.99$\pm$2.97              & 0.6565$\pm$0.1069  \\
\cmidrule{2-8}
& ADN$^\heartsuit$~\cite{adn2019_tmi}            & 22.01$\pm$3.84               & 0.8220$\pm$0.0687                & 21.61$\pm$3.69               & 0.7866$\pm$0.0760   & 21.29$\pm$3.67  & 0.7669$\pm$0.0826\\
\cmidrule{2-8}
   & CNN-MAR$^\diamondsuit$~\cite{cnnmar}                & 30.53$\pm$1.93               & 0.8148$\pm$0.0738                & 30.26$\pm$2.18              & 0.8017$\pm$0.0848      & 30.13$\pm$2.27      & 0.7956$\pm$0.0902\\
                              & DICDNet$^\diamondsuit$~\cite{wang2021dicdnet}                & 30.16$\pm$2.80               & 0.9239$\pm$0.0357               & 28.75$\pm$3.40              & 0.9061$\pm$0.0447    & 28.10$\pm$3.67 & 0.8948$\pm$0.0510 \\
                              & ACDNet$^\diamondsuit$~\cite{wang2022acdnet}                 & 29.98$\pm$3.12               & 0.9259$\pm$0.0420               & 27.57$\pm$4.38              & 0.9045$\pm$0.0608   & 26.84$\pm$4.65  & 0.8947$\pm$0.0671 \\ 
\cmidrule{2-8}
& Polyner$^\clubsuit$~\cite{wu2023unsupervised}                    & \underline{31.62$\pm$2.49}             & \underline{0.9497$\pm$0.0265}                 & \underline{31.75$\pm$2.08}               & \underline{0.9462$\pm$0.0267}  & \underline{31.37$\pm$2.28} & \textbf{0.9450$\pm$0.0284}\\
& Diner$^\clubsuit$ (Ours)                    & \textbf{33.75$\pm$1.12}               & \textbf{0.9514$\pm$0.0243}               & \textbf{33.53$\pm$1.16}               & \textbf{0.9487$\pm$0.0247}   & \textbf{33.43$\pm$1.17}
     & \textbf{0.9472$\pm$0.0254}\\
\bottomrule
\end{tabular}
\end{table*}
\begin{figure*}[t]
    \centering
    \includegraphics[width=0.95\textwidth]{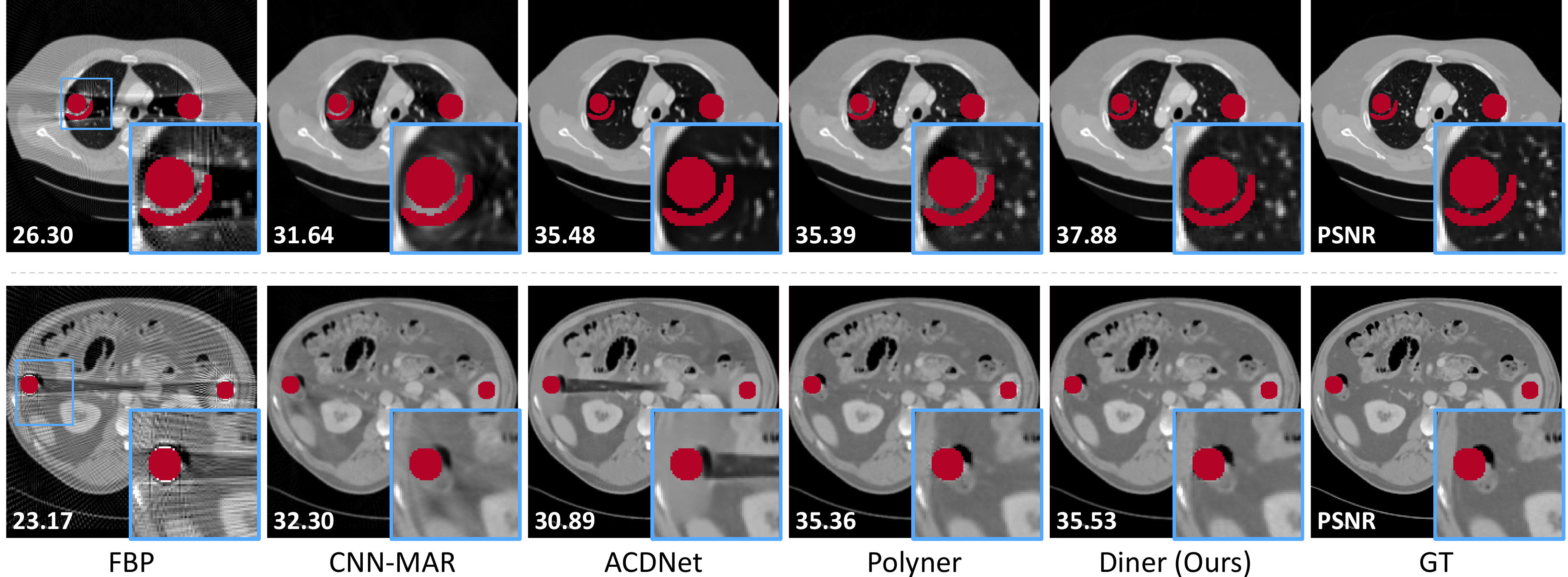}
    \caption{Qualitative results of four compared methods and our Diner on two samples (\#100 (top-two rows) and \#50 (bottom-two rows)) of the DeepLesion~\cite{deeplesion} and LIDC~\cite{LIDC} datasets. The red regions denote metals.}
    \label{fig:comparison}
\end{figure*}
\subsubsection{\textbf{Mouse Tight Real-world Dataset}}
\par To evaluate the effectiveness of our Diner model in biomedical X-ray CT settings, we scan five mouse-thigh samples containing intramedullary needles of 304 stainless steel metal on a commercial Bruker SKYSCAN 1276 micro-CT scanner, as shown in Figure~\ref{fig:fig_microCT}a. The CT acquisition protocol is shown in Table~\ref{tab:geometry}. We use the SPEKIT toolkit~\cite{punnoose2016technical} to estimate the X-ray energy spectrum of the scanner. 
\subsubsection{\textbf{Body Phantom Real-world Dataset}}
\par To test the MAR performance of the proposed method on clinical helical CT, we measure a human body phantom containing Ti metals using a commercial UIH uCT 768 scanner (Figure~\ref{fig:fig_microCT}b). To align with clinical settings, we use a standard clinical acquisition protocol, which is demonstrated in Table \ref{tab:geometry}. 
\subsection{Baselines and Evaluation Metrics}
\par We compare the proposed Diner with eight representative MAR methods belonging to four categories: \romannumeral1) Three model-based methods (FBP~\cite{fbp}, LI~\cite{kalender1987li}, and ART~\cite{art_method}); \romannumeral2) Three supervised DL methods (CNN-MAR~\cite{cnnmar}, DICDNet~\cite{wang2021dicdnet}, and ACDNet~\cite{wang2022acdnet}); \romannumeral3) One unpaired-supervised DL method (ADN~\cite{adn2019_tmi}); \romannumeral4) One unsupervised DL method (Polyner~\cite{wu2023unsupervised}). It should be noted that the supervised and unpaired-supervised DL methods are evaluated based on the pre-trained models provided by the authors to ensure a fair comparison. Among them, CNN-MAR~\cite{cnnmar} is trained on the XCOM dataset~\cite{berger2009xcom}, while DICDNet~\cite{wang2021dicdnet}, ACDNet~\cite{wang2022acdnet}, and ADN~\cite{adn2019_tmi} are trained on the DeepLesion dataset~\cite{deeplesion}. We use peak signal-to-noise ratio (PSNR) and structural similarity index measure (SSIM) as quantitative metrics.
\subsection{Comparison with Baselines on Simulation Data}
\par Table~\ref{table_comparison} shows the quantitative results. Generally, our Diner produces the best performance for all cases. On the DeepLesion dataset~\cite{deeplesion}, our unsupervised Diner not only significantly outperforms the unsupervised Polyner~\cite{wu2023unsupervised}, but also slightly performs better than the two supervised methods (DICDNet~\cite{wang2021dicdnet} and ACDNet~\cite{wang2022acdnet}). For example, the PSNR respectively improves by +1.94 dB, +1.24 dB, and +0.74 dB for the titanium metals. While on the LIDC dataset~\cite{LIDC}, the supervised ACDNet and ACDNet trained on the DeepLesion dataset~\cite{deeplesion} fail to achieve pleasant MAR performance due to the OOD problem. They even perform worse than the traditional model-based LI algorithm~\cite{kalender1987li}, such as -2.46 dB and -1.20 dB in the PSNR for the 304 stainless steels. In comparison, our Diner still yields the best MAR reconstructions, outperforming the second-best Polyner~\cite{wu2023unsupervised} by about +2.06 dB in PSNR for the 304 stainless steels. 
\par We show the qualitative results in Figure~\ref{fig:comparison}. Traditional FBP algorithm~\cite{fbp} and supervised CNN-MAR trained on the XCOM~\cite{berger2009xcom} cannot recover satisfactory results, including severe artifacts on both datasets. While ACDNet produces a clear but slightly smooth MAR reconstruction on the DeepLesion dataset~\cite{deeplesion}, its result on the LIDC dataset suffers from severe shadow artifacts. In comparison, the two unsupervised methods (Polyner~\cite{wu2023unsupervised} and our Diner) achieve robust results across the two datasets. However, the results of Polyner~\cite{wu2023unsupervised} contain slight artifacts around the metal, while our Diner produces clean and fine-detailed MAR reconstruction.
\subsection{Comparison with Baselines on Real-world Data}
\begin{figure}[t]
    \centering
    \includegraphics[width=0.475\textwidth]{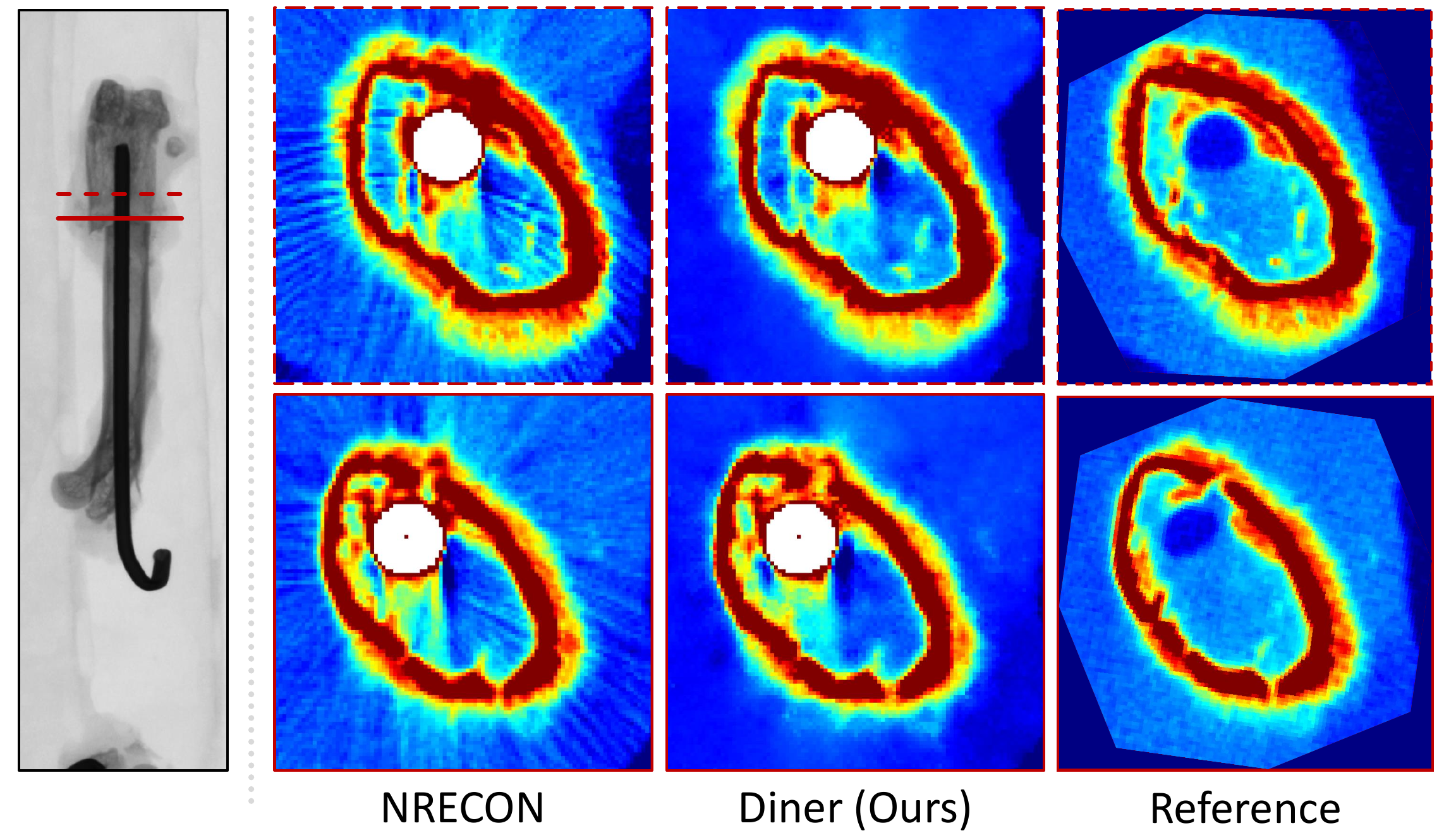}
    \caption{(Left) A sample among 2D projections for a mouse-thigh sample containing an intramedullary needle. (Right) Qualitative results of NRECON (\ie a reconstruction toolkit developed by Bruker and equipped with Bruker SKYSCAN 1276 micro-CT scanner) and our Diner on the needle-corrupted measurement. The reference image is the clear version of the mouse thigh sample after removing the intramedullary needle. The white regions denote the intramedullary needle.}
    \label{fig:real_data}
\end{figure}
\begin{figure}[t]
    \centering
    \includegraphics[width=0.475\textwidth]{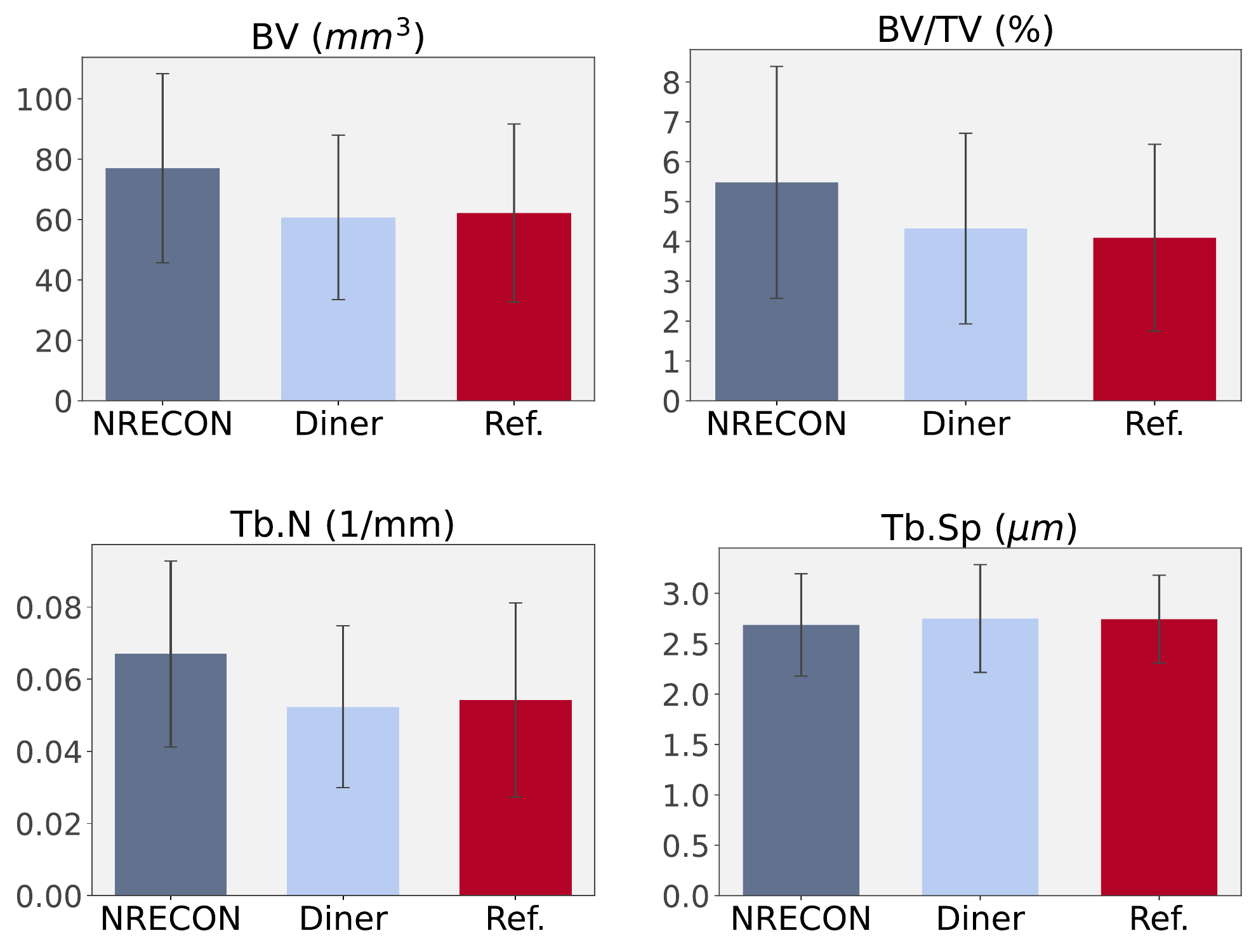}
    \caption{Qualitative results of bone analysis on the CT volumes of the five mouse-thigh samples containing an intramedullary needle. Here BV, BV/TV, Tb.N, and Tb.SN respectively denotes bone volume, percent bone volume, trabecular number, and trabecular separation. NRECON is a reconstruction toolkit developed by Bruker and equipped with Bruker SKYSCAN 1276 micro-CT scanner. Ref. is produced by NRECON on the clear measurement.}
    \label{fig:fig_real_ana}
\end{figure}
\subsubsection{\textbf{On Mouse Tight Dataset}}
\par We collect five mouse-thigh samples containing an intramedullary needle with a commercial Bruker SKYSCAN 1276 micro-CT scanner. The initial intent of the mouse model is to investigate the progress of bone-fracture healing in mice. The setup is shown in Figure~\ref{fig:fig_microCT}a. The intramedullary needle was implanted into the mouse tight one week before the CT scan. To further assess the accuracy of MAR CT reconstruction, the mouse thighs are scanned twice before and after needle removal. The reference image is the clear version of the mouse thigh sample after removing the intramedullary needle. 
\par We show the qualitative results in Figure~\ref{fig:real_data}. Compared with NRECON, a reconstruction toolkit developed by Bruker and equipped with Bruker SKYSCAN 1276 micro-CT scanner, our Diner produces a clear CT reconstruction closer to reference. Moreover, we use CTan, a bone analysis software developed by Bruker, to perform a bone analysis on these CT volumes. The results are shown in Figure~\ref{fig:fig_real_ana}. In terms of bone volume (BV), percent bone volume (BV/TV), and trabecular number (Tb.N), the reconstruction of NRECON obtain distinct higher scores than the clear reference image. This is because the metal artifacts in the CT volume are identified as high-density bone tissue by the CTan software. In comparison, our Diner effectively removes these metal artifacts and thus obtains very close scores to the reference image. 
\begin{figure}[t]
    \centering
    \includegraphics[width=0.48\textwidth]{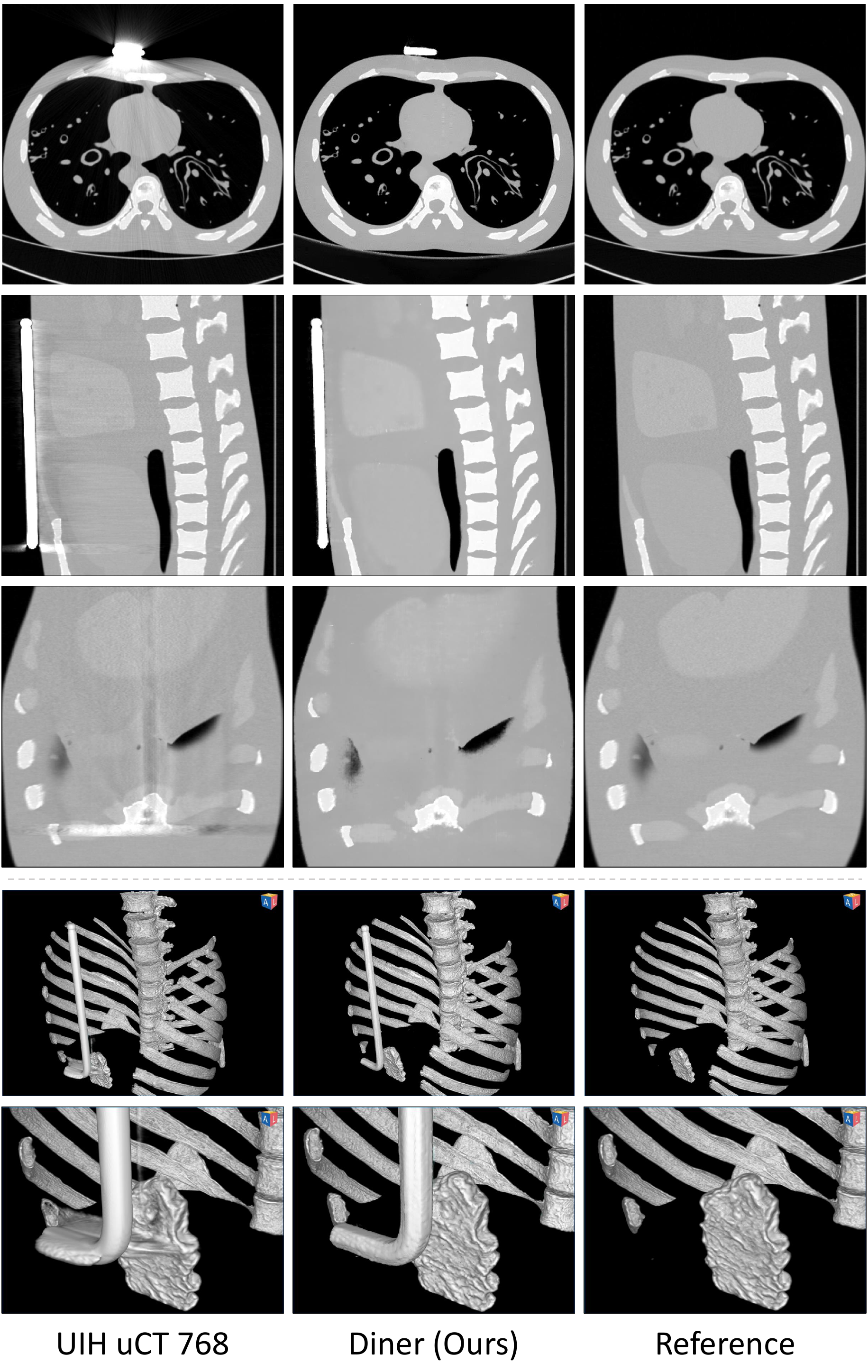}
    \caption{(Rows 1--3) Qualitative results of the reconstruction algorithm equipped with the UIH uCT 768 scanner and our Diner on the metal-corrupted measurement. The reference image is the clean reconstruction of the body phantom without the metal. (Rows 4--5) Volume rendering of the CT volumes by using the Bee DICOM Viewer software (\url{https://beedicom.com/}).}
    \label{fig:fig_phantom}
\end{figure}
\subsubsection{\textbf{On Body Phantom Dataset}}
\par We conduct scans on two versions (with and without a metal stick) of a body phantom using the UIH commercial uCT 768 scanner according to a clinical acquisition protocol (See Table~\ref{tab:geometry} for detailed parameters). The experimental setup is depicted in Figure~\ref{fig:fig_microCT}b. The qualitative results are presented in Figure~\ref{fig:fig_phantom}. Here, ``UIH uCT 768" denotes the reconstruction generated by the algorithm integrated into the clinical UIH uCT 768 scanner. Upon visual inspection of the CT slices from axial, sagittal, and coronal views (Rows 1--3), our Diner effectively reduces metal artifacts while finely preserving image details. 
\par Additionally, we utilize the Bee DICOM Viewer software for volume rendering of the CT reconstructions. These results are illustrated in Figure~\ref{fig:fig_phantom} (Rows 4--5). Also, the rendering outcome of our Diner is clear and fine-detailed, closely resembling that of the reference image. In summary, these ongoing experimental investigations conducted using the UHI uCT 768 scanner robustly demonstrate the efficacy of our Diner in clinical settings.
\begin{figure}[t]
    \centering
    \includegraphics[width=0.48\textwidth]{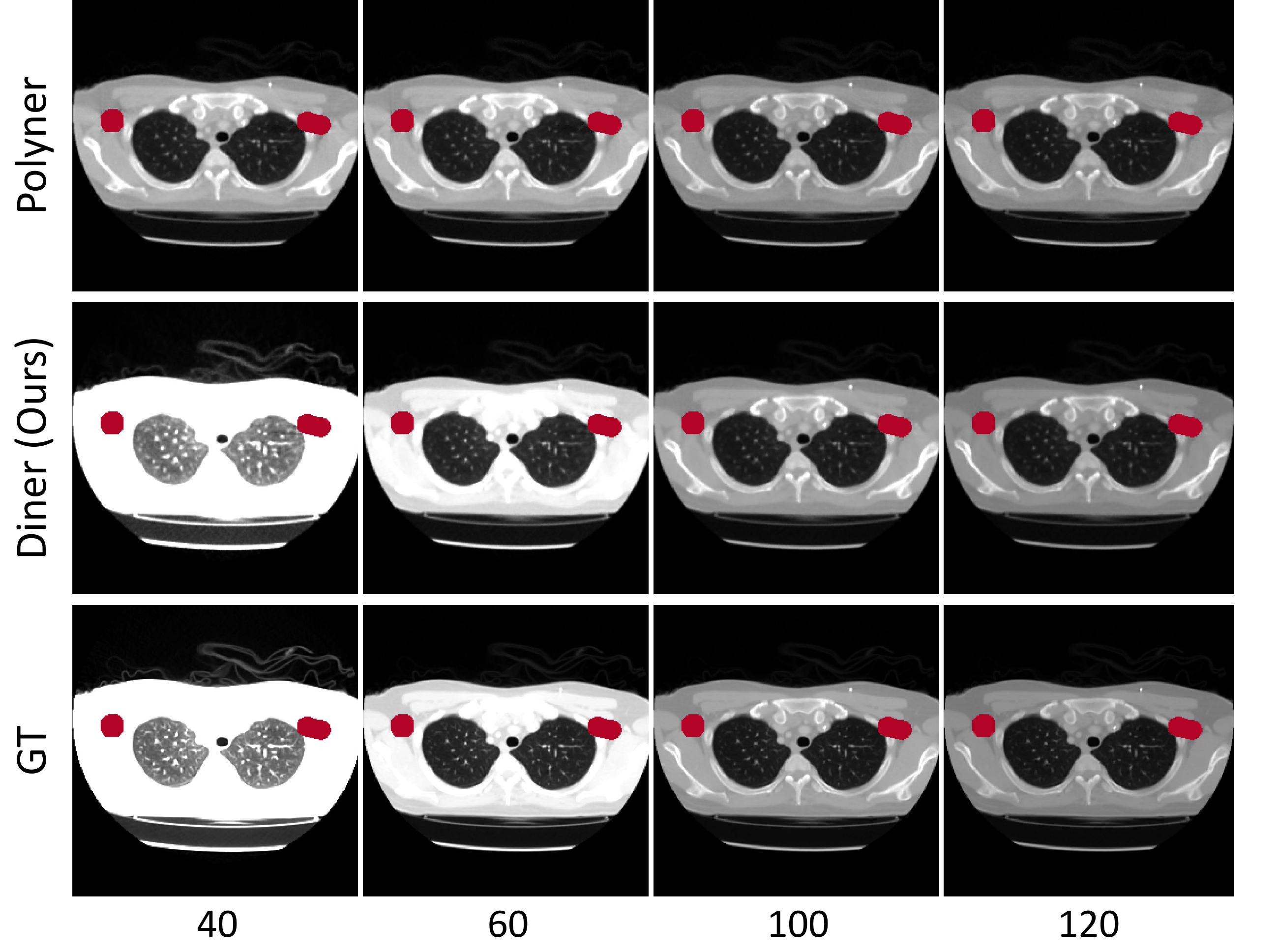}
    \caption{Qualitative results of polychromatic images at four energy levels $E=\{40, 60, 100, 120\}$ by our Diner and Polyner~\cite{wu2023unsupervised} on a sample of the DeepLesion dataset~\cite{deeplesion}. The red regions denote metals.}
    \label{fig:polychromatic_images}
\end{figure}
\begin{figure}[t]
    \centering
    \includegraphics[width=0.4\textwidth]{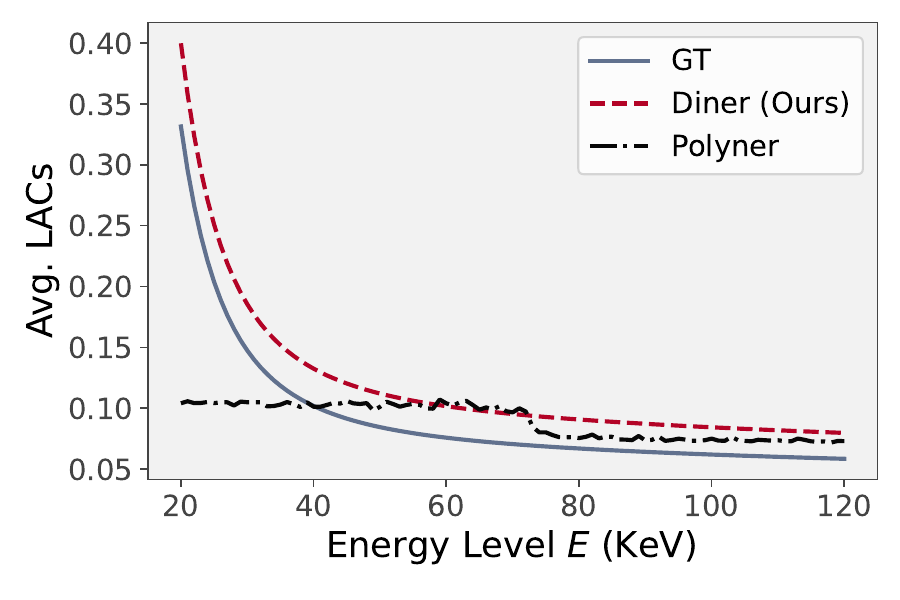}
    \caption{Average LAC curves of polychromatic images by Polyner~\cite{wu2023unsupervised} and our Diner on a sample of the DeepLesion dataset~\cite{deeplesion} over the X-ray energy levels of [20, 120] KeV.}
    \label{fig:polychromatic_curve}
\end{figure}
\subsection{Reconstruction of Polychromatic Images}
\par In Polyner~\cite{wu2023unsupervised}, the recovery of the polychromatic CT images is ill-posed on the spectrum distribution. In contrast, our Diner solves the energy-independent density map, which improves the reconstruction of the energy-dependent LACs. By conducting our LAC decomposition model (Eq. \ref{eq: lacs-density}), we can generate the LAC maps from the resolved density map. Here, we compare our Diner with Polyner for reconstructing the polychromatic images. Figure~\ref{fig:polychromatic_images} show the solved polychromatic images at four energy levels $E=\{40, 60, 100, 120\}$. From the visualization, we observe that the results of our Diner model are very close to the GT samples in terms of image structure and contrast. In comparison, the reconstructions of Polyner are satisfactory for image structure but inconsistent for image contrast, \ie spectrum distribution. 
\par We show the average LAC curve of the polychromatic images over the X-ray energy in Figure~\ref{fig:polychromatic_curve}. Polyner model produces an over-smooth spectrum distribution due to an energy-smooth regulation. In contrast, our Diner is highly consistent with GT, benefiting from the LAC decomposition model (Eq. \ref{eq: mac-lac}). To sum up, our Diner greatly outperforms our previous Polyner model for the polychromatic CT images.
\begin{figure}[t]
    \centering
    \includegraphics[width=0.48\textwidth]{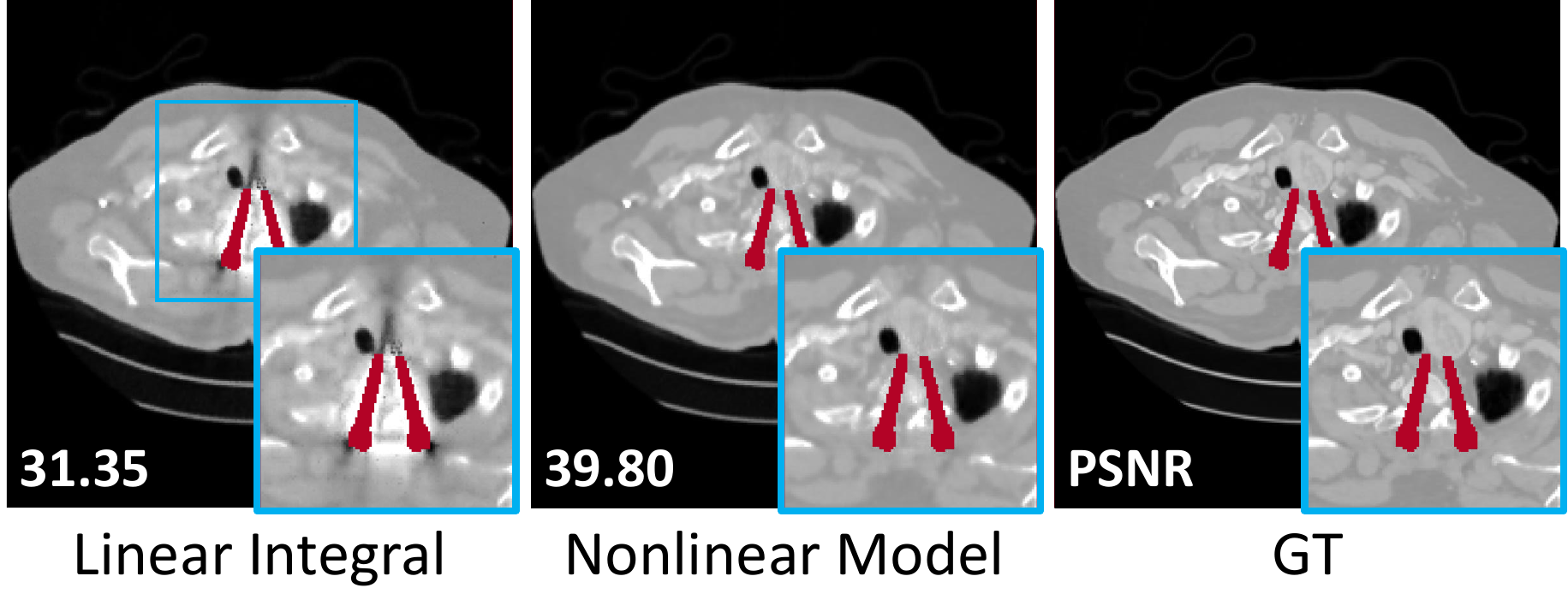}
    \caption{Qualitative results of our Diner with different forward models on a sample of the DeepLesion dataset~\cite{deeplesion}. The red regions denote metals.}
    \label{fig:ab_forward_mo9del}
\end{figure}
\begin{table}[t]
    \centering
    \caption{Quantitative results of our Diner with different forward models on the DeepLesion dataset~\cite{deeplesion}.}
    \label{table:ab_forward_model}
    \begin{tabular}{lcc}
    \toprule
     \textbf{Forward Model}    & PSNR & SSIM \\ \midrule
     Linear Integral    & 34.47$\pm$2.47 & 0.9578$\pm$0.0137\\
     Nonlinear Model $\mathcal{T}$    & \textbf{39.19$\pm$1.44} & \textbf{0.9844$\pm$0.0029} \\
     \toprule
    \end{tabular}
\end{table}
\subsection{Ablation Studies}
\subsubsection{\textbf{Influence of Forward Model}}
\par Our Diner proposes a new forward model $\mathcal{T}$ (Eq. \ref{eq: full-forward-model}) to enable the INR network to reconstruct the densities of the object. Here, we explore the influence of the forward model $\mathcal{T}$ on the MAR performance. Specifically, we replace it with the linear integral model (Eq. \ref{eq:ct-forward}) used in the existing SOTA MAR techniques~\cite{cnnmar, kalender1987li, fbp, adn2019_tmi, wang2022acdnet, wang2021dicdnet} and make all other model configurations the same for a fair comparison. Table~\ref{table:ab_forward_model} shows the quantitative results. Our proposed nonlinear model $\mathcal{T}$ produces significant performance improvements compared with the conventional linear integral, where PSNR improves by +4.72 dB. We show the qualitative results in Figure~\ref{fig:ab_forward_mo9del}. The linear model fails to tackle the shadow artifacts caused by the BHE. In comparison, the result of our nonlinear model is visually clear and close to the GT. 
\begin{figure}[t]
    \centering
    \includegraphics[width=0.48\textwidth]{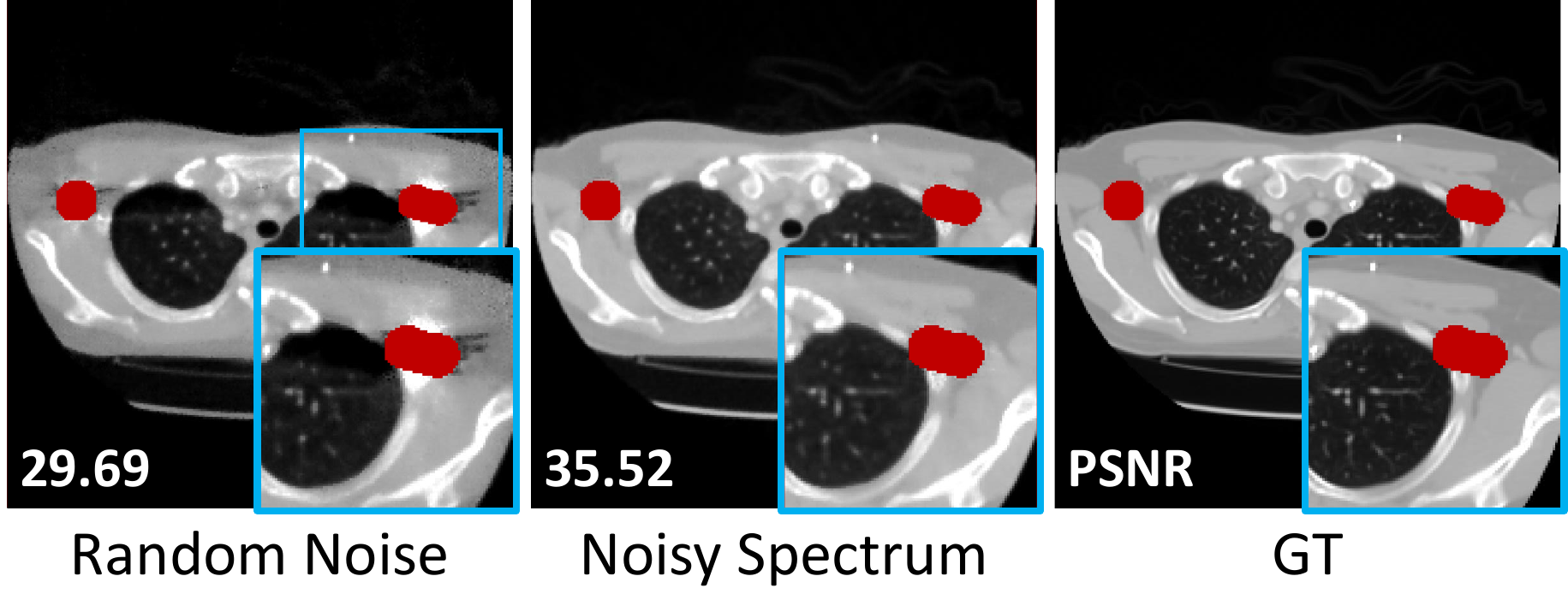}
    \caption{Qualitative results of our Diner with two types of the X-ray energy spectra (random noisy and noisy spectrum) on a sample of the DeepLesion dataset~\cite{deeplesion}. The red regions denote metals.}
    \label{fig:ab_spectrum_prior}
\end{figure}
\begin{table}[t]
    \centering
    \caption{Quantitative results of our Diner with different spectrum priors $\eta$ on the DeepLesion dataset~\cite{deeplesion}.}
    \label{table:ab_spectrum_prior}
    \begin{tabular}{ccc}
    \toprule
     \textbf{Spectrum Prior}    & PSNR & SSIM \\ \midrule
     Random Noise $\mathcal{N}(0, 1)$    & 22.45$\pm$4.64 & 0.8050$\pm$0.1529\\
     Noisy Spectrum ($3\%$)    & 35.64$\pm$1.48 & 0.9799$\pm$0.0063 \\
     Noisy Spectrum ($1.5\%$)    & 36.80$\pm$1.26 & 0.9799$\pm$0.0063 \\
     Clear Spectrum   & \textbf{39.19$\pm$1.44} & \textbf{0.9844$\pm$0.0029} \\
     \toprule
    \end{tabular}
\end{table}
\subsubsection{\textbf{Influence of Energy Spectrum Prior}}
\par In our forward model $\mathcal{T}$, the X-ray energy spectrum $\eta\in\mathbb{R}^N$ is considered as available prior knowledge. Here, we study its influence on the model performance. Three types of spectra are compared. (1) Random Noise, a vector of length $N$ sampled from the Gaussian distribution $\mathcal{N}(0, 1)$. (2) Noisy Spectrum, a spectrum corrupted by an additive noise following the Gaussian distribution $\mathcal{N}(0, \eta_\mathrm{max}\cdot c)$, where $\eta_\mathrm{max}$ denotes the maximum values of the clear spectrum, and $c$ is set to $3\%$ and $1. 5\%$ to simulate two levels of noise. (3) Clear spectrum, \ie a GT spectrum. Table~\ref{table:ab_spectrum_prior} shows the quantitative results. There are two observations. First, using the random noise as the energy spectrum causes severe degradation of the MAR performance, which decreases by 16.74 dB in terms of PSNR compared to using the clear spectrum. Second, the use of the noisy spectra decreases the performance of our Diner, but at an acceptable level. For example, the PSNR reaches 35.64 dB and 36.80 dB for the noisy spectra (1.5\% and 3\%). Figure~\ref{fig:ab_spectrum_prior} shows the qualitative results. From the visualization, we observe that the result of the noisy spectrum is significantly better than that of the random noise. In summary, the energy spectrum is required for our Diner, but it does not significantly affect the MAR performance if inaccurate.
\subsubsection{\textbf{Influence of Discretization of Energy Spectrum}}
\par
\begin{figure}[t]
    \centering
    \includegraphics[width=0.4\textwidth]{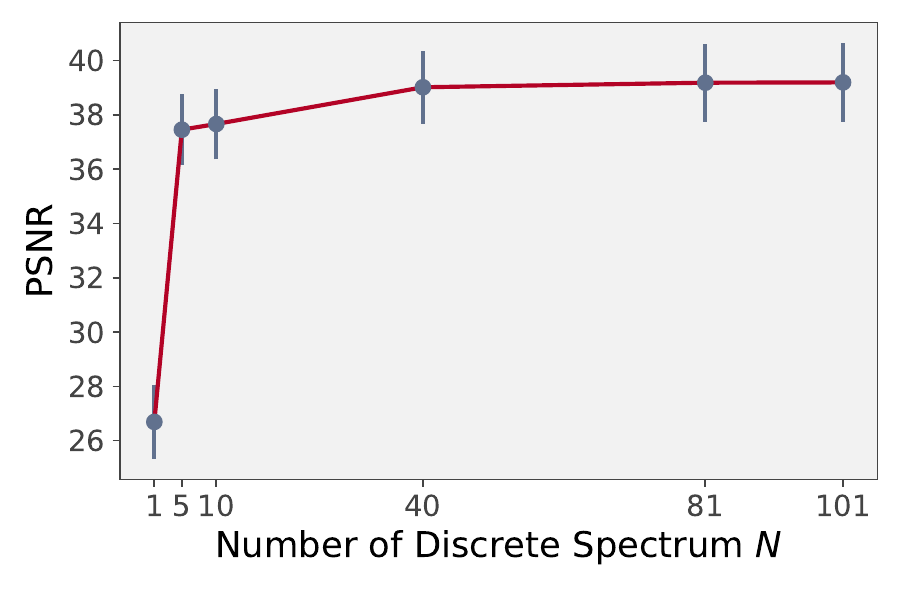}
    \caption{Performance curve of our Diner with different discrete energy spectra $N=\{101, 81, 40, 10, 5, 1\}$ on the DeepLesion dataset~\cite{deeplesion}. Note that $N=1$ means the degradation of the proposed forward model $\mathcal{T}$ towards the linear integral model (Eq. \ref{eq. measurement define}).}
    \label{fig:ab_discrete_energy_spectrum_curve}
\end{figure}

\begin{figure}[t]
    \centering
    \includegraphics[width=0.48\textwidth]{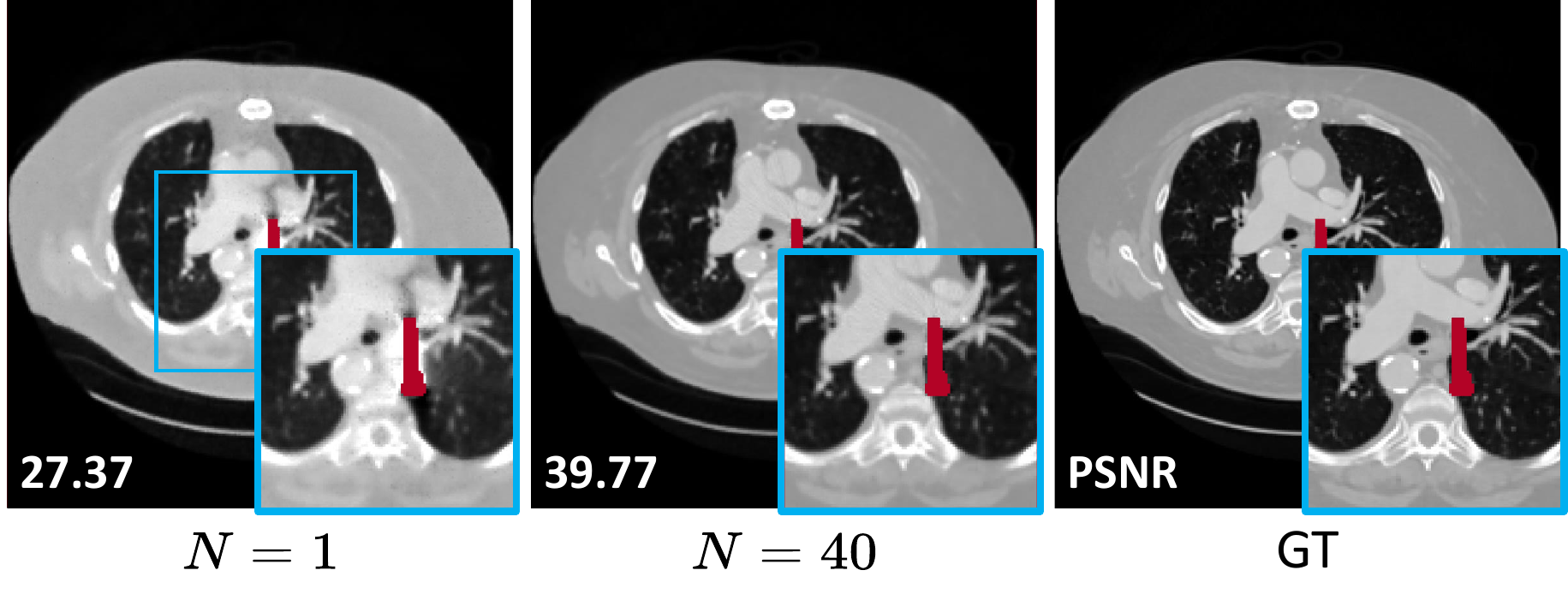}
    \caption{Qualitative results of our Diner with two discrete energy spectra $N=\{1, 40\}$ on a sample of the DeepLesion dataset~\cite{deeplesion}. Note $N=1$ means the degradation of the proposed forward model $\mathcal{T}$ towards the linear integral model (Eq. \ref{eq. measurement define}). The red regions denote metals.}
    \label{fig:ab_discrete_energy_spectrum}
\end{figure}

 The energy of a polychromatic X-ray physically covers a continuous range. However, due to limitations of physical instruments and numerical methods, measuring or estimating energy spectra are often based on discrete forms~\cite{punnoose2016technical, boone1997accurate, poludniowski2009spekcalc}. Our forward model $\mathcal{T}$ also uses the discrete energy spectrum (Eq. \ref{eq: dis-forward-model}). Here, we discuss how this discretization affects the MAR performance. We set the energy range to [20, 120] KeV and then uniformly sample $N=\{1, 5, 10, 40, 81, 101\}$ energy levels from the range. Figure~\ref{fig:ab_discrete_energy_spectrum_curve} shows the performance curve of our Diner over the discrete energy spectrum. From $N=5$ to $N=101$, the model performance remains excellent and robust (the change of the PSNR is less than 2 dB). When $N$ is set to 1, our Diner fails to produce good results (PSNR is only about 27 dB). This is because the forward model $\mathcal{T}$ degrades to a linear integral model (Eq. \ref{eq:ct-forward}) when $N=1$, in which the energy-induced BHE cannot be handled. The qualitative results are shown in Figure~\ref{fig:ab_discrete_energy_spectrum}. Visually, the MAR reconstruction is satisfactory in both local details and global structures when $N=40$, while it contains several metal artifacts when $N=1$. In summary, the MAR performance of our Diner is not affected by the discretization of the energy spectrum. 
\subsection{Failure Cases}
\par
\begin{figure}[t]
    \centering
    \includegraphics[width=0.48\textwidth]{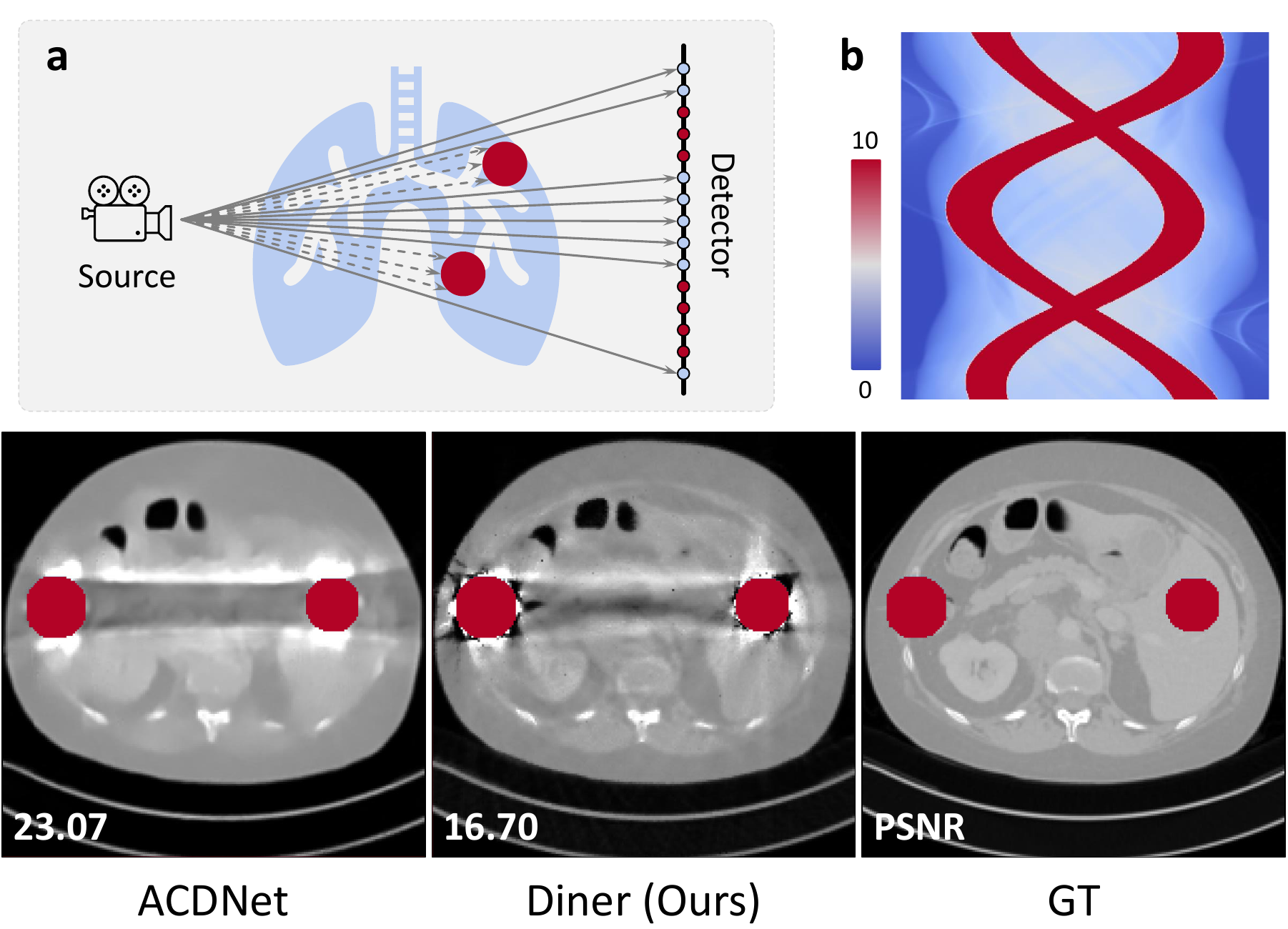}
    \caption{A failure case of our Diner for ultra-high absorption metals. (\textbf{a}) A schema of the photon starvation effect, (\textbf{b}) A simulated gold-corrupted CT measurement on the DeepLesion~\cite{deeplesion} dataset, and (\textbf{bottom row}) Qualitative and quantitative results of ACDNet~\cite{wang2022acdnet} and our Diner for the gold-corrupted measurement. The red regions denote metals.}
    \label{fig:failure_case}
\end{figure}

 Our Diner proposes a novel nonlinear forward model $\mathcal{T}$ (Eq. \ref{eq: full-forward-model}) to account for the complex BHE. However, there is another challenge when imaging objects containing ultra-high absorption metals, such as gold. As shown in Figure~\ref{fig:failure_case}a, the X-rays hitting the ultra-high absorption metals are almost absorbed, resulting in the loss of valid measurement signals at the corresponding locations of the detectors. This phenomenon is called the photon starvation effect. Figure~\ref{fig:failure_case}b shows a simulated gold-corrupted CT measurement, where the red regions represent the invalid signals and are background noise. However, our Diner considers the entire measurements as valid signals and uses them to optimize the INR network, causing the gradient error propagation. We show the MAR reconstructions by the SOTA-supervised ACDNet~\cite{wang2022acdnet} and our Diner for the gold-corrupted CT measurement in Figure~\ref{fig:failure_case} (bottom row). Visually, the results of both techniques contain severe shadow artifacts and are unacceptable image quality. In summary, our Diner fails to handle the photon starvation effect caused by the ultra-high absorption metals. Nevertheless, we would like to emphasize that the common medical metal implants (\eg titanium, chromium, and 304 stainless steel) do not absorb X-rays as well as gold. Therefore, the proposed approach works for most clinical scenarios.

\section{Conclusion and Limitation}
\par We present Diner, a novel reconstruction technique based on neural representation, to address the challenging MAR problem from a nonlinear perspective. The proposed Diner is fully unsupervised and thus does not require any external training data, significantly enabling its usefulness in a wide of clinical scenarios. Our Diner learns a neural representation of the energy-independent densities of observed objects to handle the nonlinear BHE fundamentally. Extensive comparisons with popular MAR approaches and ablation studies confirm the effectiveness and reliability of the proposed method. 
\par Although our Diner shows great potential for the MAR problem, several underlying limitations exist. For example, our Diner requires basic CT acquisition information, such as acquisition geometry and raw measurement data. However, the information might be inaccessible on clinical CT scanners due to commercial privacy. 

\bibliography{ref}
\bibliographystyle{ieeetr}

\end{document}